# Estimating the severity of dental and oral problems via sentiment classification over clinical reports

**Sare Mahdavifar**[1], **Seyed Mostafa Fakhrahmad**[2*], **Elham Ansarifard**[3]

[1]Dept. of Computer Science and Engineering and IT, Shiraz University, Shiraz, Iran, sare.mahdavifar@hafez.shirazu.ac.ir
[2]Dept. of Computer Science and Engineering and IT, Shiraz University, Shiraz, Iran, fakhrahmad@shirazu.ac.ir
[3]Dept. of Prosthodontics, School of Dentistry, Shiraz University of Medical Sciences, Shiraz, Iran, ansarifard@sums.ac.ir

*Abstract*—Analyzing authors' sentiments in texts as a technique for identifying text polarity can be practical and useful in various fields, including medicine and dentistry. Currently, due to factors such as patients' limited knowledge about their condition, difficulties in accessing specialist doctors, or fear of illness, particularly in pandemic conditions, there might be a delay between receiving a radiology report and consulting a doctor. In some cases, this delay can pose significant risks to the patient, making timely decision-making crucial. Having an automatic system that can inform patients about the deterioration of their condition by analyzing the text of radiology reports could greatly impact timely decision-making.

In this study, a dataset comprising 1,134 cone-beam computed tomography (CBCT) photo reports was collected from the Shiraz University of Medical Sciences. Each case was examined, and an expert labeled a severity level for the patient's condition on each document. After preprocessing all the text data, a deep learning model based on Convolutional Neural Network (CNN) and Long Short-Term Memory (LSTM) network architecture, known as CNN-LSTM, was developed to detect the severity level of the patient's problem based on sentiment analysis in the radiologist's report. The model's performance was evaluated on two datasets, each with two and four classes, in both imbalanced and balanced scenarios. Finally, to demonstrate the effectiveness of our model, we compared its performance with that of other classification models. The results, along with one-way ANOVA and Tukey's test, indicated that our proposed model (CNN-LSTM) performed the best according to precision, recall, and f-measure criteria. This suggests that it can be a reliable model for estimating the severity of oral and dental diseases, thereby assisting patients.

*Keywords—Sentiment Classification, Deep Learning, Document Classification, Machine Learning.*

## I. INTRODUCTION

CBCT radiography is a type of dental imaging that captures three-dimensional images of the teeth, as well as all the bones and soft tissues within the mouth. Analyzing three-dimensional CBCT images has become an indispensable procedure for the diagnosis and treatment planning of orthodontic patients [1]. The applications of CBCT radiography are diverse, including informing attending physicians of hidden teeth, diagnosing joint disorders, precisely placing implants, examining the condition of the jaw, sinuses, and nerve canals, diagnosing tumors in the jaw and mouth, assessing bone structure, identifying the source and origin of pain, and guiding surgical interventions. As a result, CBCT imaging, unlike other methods, can reveal a broad spectrum of oral, jaw, and facial issues. Some of these issues require urgent attention, such as the presence of malignant tumors in the mouth and jaw areas, necessitating swift treatment to prevent endangering the person's life. In other cases, there are significant issues that require follow-up and treatment, but the urgency may not be as high, and a delay may not pose a serious risk. The third scenario arises when a problem is identified in CBCT images, which an individual can choose to correct, but there is no compulsion or necessity to do so. For instance, in the upper jaw, the bone depth near the nerve canals and sinus cavities may be low. If the individual decides to implant in that area later on, a sinus lift would be necessary to create more depth and distance. However, since this is not considered a problem or a disease, and the person may choose not to undergo implantation, addressing this matter is not particularly urgent. The fourth scenario is when conditions are entirely normal, and there is no problem requiring treatment or correction. Fig. 1 demonstrates a sample of CBCT images.

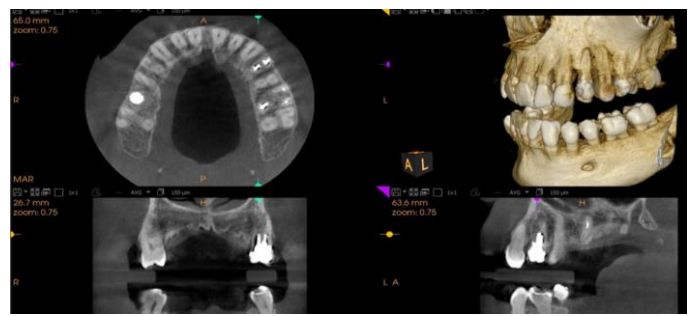

Figure 1. Sample of CBCT images

Factors such as patients' lack of knowledge about their conditions, the unavailability of specialist doctors, unfavorable financial conditions, work-related issues, or the fear associated with being in a medical environment (especially during a pandemic) can contribute to a delay in receiving a radiology report and consulting a doctor. In some cases, this time delay, as seen in the first of the four scenarios of radiology report results, can pose significant and irreparable risks for the patient. In such situations, designing a system capable of accurately diagnosing the severity of the patient's problem by examining and analyzing the radiology report becomes crucial. This acquired knowledge holds significance at various stages of a patient's life, particularly in aiding physicians in assessing a patient's health progress during treatment. The extraction of knowledge from these summaries enhances the evaluation of treatment quality, providing benefits to both patients and healthcare facilities [2, 3, 4]. Providing timely warnings about the need for follow-up and prompt treatment can shield the patient from the risks associated with procrastination and delay. On the other hand, if there is no acute problem, informing the patient of this condition can help alleviate anxiety before their doctor's visit.

---
* Corresponding author

To achieve the stated goal, the approach of sentiment analysis in the text can be employed. This approach allows the severity and seriousness of the problem to be reported based on the way the radiologist expresses the issue in the text. Sentiment analysis is a branch of natural language processing (NLP), which, in simple terms, involves examining the point of view and opinion expressed about a specific topic in a text.

Sentiment analysis techniques are designed to determine the expressed sentiment regarding a particular subject or the overall contextual polarity of a document. Emerging from the realm of web mining, the advancement of sentiment analysis approaches frequently focuses on analyzing highly subjective texts, notably customer reviews [5, 6, 7]. This operation finds applications in various fields. For instance, in the commercial sector, it can be used to analyze customer opinions about a particular product. Similarly, it has valuable applications in the field of medicine and health. There is a wealth of healthcare information available online, such as personal blogs, social media, and websites ranking medical topics, and extracting this information may not be straightforward to obtain. Therefore, sentiment analysis offers numerous benefits, including leveraging medical information to achieve optimal results for improving healthcare quality. Approaches in sentiment analysis can be categorized based on the techniques used, the structure of the dataset, the level of ranking, and other factors.

In recent times, sentiment analysis has made inroads into the realm of medicine. Leveraging the advances in big data, medical professionals are harnessing this opportunity to diagnose and analyze emotions using available data, ultimately improving medical services. This approach departs from traditional systems that rely on patient questionnaires for initial diagnosis. NLP enables the analysis of the vast patient dataset, which is then structured into a training dataset to educate an automated system. This research is dedicated to distinguishing high-risk patients from low-risk patients and providing crucial assistance, particularly during challenging situations such as the Pandemic. We meticulously annotated a dataset comprising reports of patients' CBCT images. Following the preprocessing phase, we employed word embedding to extract features for training the CNN-LSTM model. This comprehensive analysis was conducted on two datasets, one with two classes and another with four distinct classes.

In the final stage, we compared the performance of our suggested model with other ML algorithms such as Multinomial Naive Bayes (MNB), Logistic Regression (LR), Linear Support Vector Classifier (LSVC), Random Forest (RF), Multilayer Perceptron (MLP), Decision Tree (DT), Support Vector Machines (SVM). Notably, the CNN-LSTM exhibited a significantly higher accuracy compared to the other models, showcasing its effectiveness in accurately segregating high-risk and low-risk patients.

## II. RELATED WORK

Categorizing text in the context of health is regarded as a distinct instance within text classification. Machine learning algorithms within NLP have proven effective in this domain. For instance, tasks such as classifying patient record notes have seen successful applications of algorithms like Support Vector Machines and Latent Dirichlet Allocation [8, 9].

Sentiment analysis in machine learning primarily relies on supervised learning and ensemble techniques. In the supervised learning approach, a dataset containing labeled instances of tweets is utilized to train a machine learning model using classification algorithms like SVM, Bayesian classifier, and Entropy classifier [10, 11, 12, 13, 14]. This training process classifies tweets into sentiment categories such as positive, neutral, and negative. Subsequently, the trained model is employed to predict the sentiment of new tweets. One notable drawback of the machine learning approach is its dependence on the creation of a substantial labeled training dataset, as the model's performance is directly influenced by the quality and quantity of the dataset [15, 16].

An additional study was carried out to examine Persian tweets to gain insights into the sentiments of the Iranian population regarding domestically produced versus foreign-made COVID-19 vaccines. Utilizing a dataset comprising over 800,000 tweets spanning from April to September 2021, they employed a deep-learning sentiment analysis model based on CNN and LSTM network architecture, known as CNN-LSTM. The findings revealed a subtle contrast in Persian sentiments toward domestic and imported vaccines, with foreign vaccines receiving more positive perceptions [17, 18].

Another study was conducted on machine learning in the field of tweet classification. In this research, a machine learning algorithm based on optimization is proposed for classifying tweets into different categories. The proposed model is evaluated in three stages including data preprocessing, and feature extraction using an optimization method, and finally, a training set is updated to classify into three different classes: positive, negative, and neutral. The proposed algorithm has a maximum accuracy of 89.47% compared to other machine learning algorithms, is faster, and reduces overall data processing time which is more suitable for larger datasets [19].

Poornima and his colleagues [20] compared the performance of different machine learning algorithms in sentiment analysis on Twitter data. The proposed method uses the frequency of terms to find the sentiment polarity of a sentence. The study compares the performance of MNB, SVM, and LR algorithms in sentence classification for sentiment analysis on Twitter data. The methodology involves the standard process of Twitter data preprocessing, feature extraction, model training, and testing. LR stands out with the highest accuracy, particularly when utilizing n-gram and bigram models. This research [21] examined various techniques and approaches in the field of sentiment analysis. The proposed algorithm includes three steps: data filtering, model training, and model testing. The algorithm uses a feature selection method, and sentiment analysis is performed using three machine learning algorithms: NB, SVM, and KNN. The results showed that the SVM algorithm outperformed the other two algorithms.

Another study [22] combined medical notes and movie review datasets to train a sentiment analysis algorithm using Word2Vec and simple deep learning. Despite challenges such as the lack of explicit annotation in medical notes and complex medical concepts, the model was trained on 80% of 50,299 sentences, 50,000 of which were from movie review data and 299 were manually annotated medical notes from the Stanford Cancer Research Center. The evaluation revealed lower accuracy on the medical notes dataset, possibly due to annotation challenges, while the movie review dataset showed better accuracy.

We introduced a DL algorithm designed to identify high-risk patients from low-risk ones, utilizing a newly acquired dataset collected by our team and annotated in collaboration with a dentistry expert. This collaboration, combining technology and healthcare, results in a system capable of meeting patients' requirements, particularly in challenging situations like a pandemic.

### III. EXPERIMENTS

#### 1 Dataset

In this study, we used the dental radiology database provided by Shiraz University of Medical Sciences as our primary data source. From this extensive database, a total of 1,134 patient records were carefully extracted to form the basis of our research. The process of data extraction involved categorizing the status of each patient according to our specific requirements. It's worth noting that clinical records often contain introductory information that does not contribute to the practical insights needed for our research as shown in Fig. 2. Moreover, given the absence of existing annotations, we enlisted the expertise of trained professionals who performed annotations manually on the selected reports, ensuring diversity in the selection to capture various clinical situations accurately. Subsequently, the annotated reports were organized into four distinct groups regarding to what was mentioned in the introduction, each labeled as 1, 2, 3, or 4, based on their content and characteristics. Table 1 illustrates a sample of each class with its label. Notably, due to the variance in sample sizes among these groups, we employed a random over-sampling technique to mitigate potential bias and ensure a balanced representation of data across all categories.

Figure 2. Sample of patients report

Table 1. Example of patients status from different labels

| Patient's report | Label |
|---|---|
| CBCT image of the proposed area was prepared for the patient based on your order. As you see on the images:<br>There is a well-defined mixed lesion with **thick sclerotic border** in left side of mandible which is extended from #n to #, .The lesion caused considerable **expansion of lingual cortical plate** and thinning of this **cortex and buccally displacement of IAN**.<br>Based on CBCT data, the most probable diagnoses are:<br>**Ossifying fibroma, Chondrosarcoma** | 1 |
| CBCT image of the proposed area was prepared for the patient based on your order. As you see based on the images:<br>The super **neumary tooth (SN)** is **oriented horizontally**. It is located palatally to #q tooth. It caused **displacement of tooth #q**, blunting of #q and **loss of continuity of palatal cortex**. SN has **involved the incisive foramen.** | 2 |
| CBCT of the #,tooth was prepared based on your order. As you see on the images:<br>**The #, tooth is horizontally oriented**. Direct contact (with preserving cortex) is noticed between the **IAN** canal and #,.Thinning of **lingual cortical plate is evident.** | 3 |
| Thanks for your consult; CBCT of the mandible was performed for the patient based on your request. Measurements for **implant placement** were done for the patient and are presented on the CBCT sheets. Note: *There is an **enostosis like bony structure** in the left side of the mandible in the molar region. | 4 |

#### 2 Methodology

In this section, we present an overview of our approach, as depicted in Fig. 3. Our proposed method consists of three distinct phases: data extraction and preprocessing, training the CNN-LSTM model, and testing this model.

The collected data are classified into four classes based on the severity of the patient's conditions, labeled 1, 2, 3, and 4.

The experiments were conducted on two datasets in both balanced and imbalanced modes: one with the dataset containing four categories, and the other with samples labeled 1 and 2, considered emergency patients, merged into a single class labeled 1, and samples labeled 3 and 4 merged into another class labeled 2 as non-emergency patients. We made this decision due to the proximity of the severity rates in groups 1 and 2, as well as groups 3 and 4. In this mode, the desired model is trained on a dataset with two categories.

First, we balanced the categories using the Oversampling method to prevent bias in the results. The dataset may contain numerous unwanted symbols and numbers that need correction or removal to enhance efficiency. Therefore, this process involves the removal of unwanted symbols and numbers. The objective of this task is to assess the performance of our proposed deep learning algorithm in analyzing patients' statuses. The applied method for addressing this problem comprises three steps:
- Data extraction and preprocessing
- Model training
- Model testing

### A. Data extraction and preprocessing

Information extraction (IE) is an NLP technique that entails the automatic extraction of structured information from unstructured text. The objective of information extraction is to transform textual data into a more organized and usable format. This process usually includes identifying specific entities, relationships, and attributes within a given text. As depicted in Fig. 2, each patient's record contains various information and attributes, of which only the text report of the CBCT image is required.

Data preprocessing involves preparing raw data before constructing Machine Learning models. In this step, all clinical reports with various labels undergo preprocessing. The dataset may contain numerous undesirable symbols and numbers that require correction or removal to enhance performance. Therefore, this process entails the removal of all symbols and numbers. For example, in the patient's report labeled as 3 in Table 1, all such symbols (#,.:*()) will be eliminated.

### B. Model training

Fetching the dataset from the file is followed by the extraction of all corresponding words (feature words) through word embedding, which represents words as vectors in a multi-dimensional space, capturing semantic relationships between them. These embeddings enhance the understanding of word meanings in natural language processing tasks, contributing to improved performance across various applications.

This process involves the removal of symbols and numbers from the CBCT's text report, and preparing the dataset for training by the CNN-LSTM model. The datasets are categorized into labels: "1" for patients in extremely critical circumstances, "2" for patients in a critical state, "3" for patients with non-severe conditions, and "4" for patients with no identified risk. Subsequently, our model is trained using the collected words, with 80% of the data allocated to each fold. To enhance training effectiveness, data shuffling is performed using a random seed. The labeled datasets are divided into 80-20% splits for training and testing, respectively. The training data is then input into the CNN-LSTM model, operating at the document level

### C. Model testing

In this phase, our CNN-LSTM model predicts the class label for each patient's report using the test dataset, which constitutes 20% of the entire dataset. This step enables a comprehensive assessment of the model's effectiveness.

The process begins with data preprocessing, encompassing the removal of symbols and numerical values from the input data. Subsequently, the carefully prepared dataset is fed into the pre-trained model for predictive analysis. After the prediction, the model's accuracy, precision, recall, and ROC-AUC metrics are meticulously evaluated to gauge its performance.

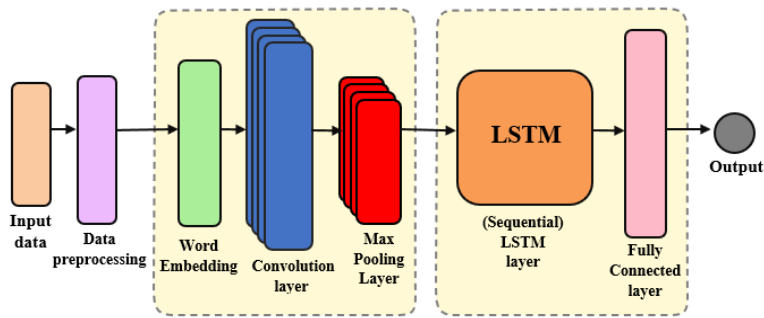

Figure 3. Scheme of our purposed method

### IV. RESULTS AND EVALUATIONS

#### 1 Evaluation metrics

We used different evaluation metrics to investigate the performance of models, including accuracy, precision, recall, F1-score, and ROC-AUC. All of these metrics are calculated based on the following parameters:
- True Positives (TP): The count of reports with an expected label correctly classified.
- True Negatives (TN): The count of reports with an unexpected label is accurately classified.
- False Positives (FP): The count of reports incorrectly classified as expected results.
- False Negatives (FN): The count of reports incorrectly classified as unexpected results.

**Accuracy** is described as the ratio of correct outcomes to the total cases evaluated, and it is applicable to both binary and multiclass classification challenges.

$$Accuracy = \frac{TP+TN}{TP+FP+TN+FN} \quad (1)$$

**Precision** of a classification is defined by the proportion of accurately classified exact and expected observations out of all specific observations.

$$Precision = \frac{TP}{TP+FP} \quad (2)$$

**Recall** is an assessment of the effectiveness in classifying positive observations. It is expressed as the ratio of True Positives to the total number of positive observations.

$$Recall = \frac{TP}{TP+FN} \quad (3)$$

**F1-Score** combines the precision and recall scores of a model.

$$F1\_Score = \frac{2*Precision*Recall}{Precision+Recall} \quad (4)$$

**AUC-ROC** evaluates how well a binary classification model can distinguish between the two classes.

## 2 Results

In this section, we assessed the effectiveness of our CNN-LSTM model by comparing it with other ML models, including MNB, LR, LSVC, RF, MLP, DT, and SVM, across both datasets in balanced and imbalanced mode. Specific parameters were employed in training the deep learning model, utilizing the Categorical Cross-Entropy loss function and the Adam algorithm for optimization. The neural networks underwent training for 10 epochs with a batch size of 32, and K-fold cross-validation (k=5) was implemented to enhance the model's performance. Dataset splitting was carried out at an 80:20 ratio, allocating 80% for training the models and reserving 20% for evaluating their performance. Moreover, All ML models are trained using the transformed data through TF-IDF, with 80% of the data assigned to each fold. To optimize training effectiveness, data shuffling is carried out using a random seed. The labeled datasets are partitioned into 80-20% splits for training and testing, respectively. The training data is subsequently fed into a variety of models, encompassing MNB, LR, LSVC, RF, MLP, DT, and SVM, all operating at the document level.

Table 2 displays the outcomes of models on an imbalanced dataset containing four classes. LSVC exhibits superior performance compared to other machine learning methods, achieving an accuracy of 69.6%. CNN-LSTM outperforms all models, achieving the highest accuracy at 90.6%. Consequently, the overall performance of deep learning surpasses that of machine learning algorithms on the imbalanced dataset with four distinct classes.

Table 3 presents the results of models on a balanced dataset comprising four classes. SVM demonstrates exceptional performance, outperforming other machine learning methods with an accuracy of 90.7%. CNN-LSTM surpasses all models, achieving the highest accuracy at 97.3%. Thus, the overall efficacy of deep learning exceeds that of machine learning algorithms on the balanced dataset with four distinct classes. Furthermore, the results indicate an improvement in the performance of all models in the balanced dataset.

Table 4 shows the results of model evaluations on an imbalanced dataset encompassing two classes. LSVC demonstrates outstanding performance, outshining other machine learning methods with an accuracy, precision, and recall of 80.3%, an f-measure of 80.2%, and an ROC-AUC of 87.7%. CNN-LSTM outperforms all models, attaining the highest accuracy, precision, and recall, scoring 95.3%, 95.1%, and 97.7% for f-measure and ROC-AUC, respectively. Consequently, the overall effectiveness of deep learning exceeds that of machine learning algorithms on the imbalanced dataset with two distinct classes.

The results of model evaluations on a balanced dataset with two classes are presented in Table 5. MLP exhibits exceptional performance, outperforming other machine learning methods with an accuracy of 93.0%, a recall of 93.2%, and an f-measure of 96.3%. CNN-LSTM excels among all models, achieving the highest accuracy, recall, and f-measure, scoring 98.0%, 98.1%, and 99.7% for precision and ROC-AUC, sequentially. Consequently, the overall efficacy of deep learning surpasses that of machine learning algorithms on the balanced dataset with two distinct classes. Moreover, the outcomes demonstrate enhanced performance for all models in the balanced dataset.

Examining Tables 4 and 5 reveals that the models demonstrate improved performance in predicting two classes compared to four classes, both in imbalanced and balanced datasets.

Table 2. The performance of models for imbalanced dataset with four classes

| Classifiers\ Evaluation metrics | Accuracy |
|---|---|
| NB | 59.8 |
| LR | 65.3 |
| **LSVC** | **69.9** |
| MLP | 66.0 |
| SVM | 64.7 |
| DT | 57.9 |
| RF | 63.3 |
| **CNN-LSTM** | **90.6** |

Table 3. The performance of models for balanced dataset with four classes

| Classifiers\ Evaluation metrics | Accuracy |
|---|---|
| NB | 75.5 |
| LR | 84.0 |
| LSVC | 88.2 |
| MLP | 90.4 |
| **SVM** | **90.7** |
| DT | 85.5 |
| RF | 90.6 |
| **CNN-LSTM** | **97.3** |

Table 4. The performance of models for imbalanced dataset with two classes

| Classifiers\ Evaluation metrics | Accuracy | Precision | Recall | F-measure | ROC-AUC |
|---|---|---|---|---|---|
| NB | 75.2 | 76.2 | 75.2 | 73.7 | 85.5 |
| LR | 77.4 | 77.3 | 77.4 | 76.8 | 87.0 |
| **LSVC** | **80.3** | **80.3** | **80.3** | **80.2** | **87.7** |
| MLP | 79.1 | 79.2 | 79.1 | 79.1 | 86.2 |
| SVM | 78.4 | 78.2 | 78.4 | 78.1 | 87.1 |
| DT | 71.8 | 71.6 | 71.8 | 71.6 | 70.1 |
| RF | 78.7 | 78.5 | 78.7 | 78.4 | 86.6 |
| **CNN-LSTM** | **95.3** | **95.3** | **95.3** | **95.1** | **97.7** |

Table 5. The performance of models for balanced dataset with two clsses

| Classifiers\ Evaluation metrics | Accuracy | Precision | Recall | F-measure | ROC-AUC |
|---|---|---|---|---|---|
| NB | 86.3 | 86.9 | 86.3 | 86.3 | 94.3 |
| LR | 88.1 | 88.4 | 88.1 | 88.0 | 95.6 |
| LSVC | 91.6 | 91.8 | 91.6 | 91.6 | 97.0 |
| **MLP** | **93.0** | **93.2** | **93.0** | **93.0** | **96.3** |
| SVM | 91.8 | 92.0 | 91.8 | 91.8 | 97.7 |
| DT | 88.0 | 88.3 | 88.0 | 88.0 | 88.0 |
| RF | 90.3 | 90.5 | 90.3 | 90.3 | 97.9 |
| **CNN-LSTM** | **98.0** | **98.1** | **98.0** | **98.0** | **99.7** |

Additionally, two statistical tests, namely, the one-way ANOVA and Tukey's HSD test, were conducted to identify significant differences in the provided comparisons. The one-way ANOVA test aims to detect an overall significant difference between the results, while Tukey's HSD test facilitates pairwise comparisons to identify which specific pairs exhibit significant differences. According to the results of Tukey's HSD test, the MLP, SVM, LSVC, and CNN-LSTM methods demonstrate significantly better performance compared to the others. Specifically, as Table 6 and Fig. 4 reveal, the Tukey HSD test indicates a statistically significant difference in the performance of the CNN-LSTM method compared to all other methods.

Nevertheless, the one-way ANOVA test reveals a significant difference across the results overall, suggesting that at least one of the methods exhibits statistically significant superiority.

Table 6. The results of Tukey's HSD test

| Pair | Difference | SE | Critical mean | p-value |
|---|---|---|---|---|
| (CNN_LSTM) - LSVC | 5.64 | 0.8737 | 3.5351 | **0.001629** |
| (CNN_LSTM) - MLP | 4.66 | 0.8737 | 3.5351 | **0.008158** |
| (CNN_LSTM) - SVM | 5.34 | 0.8737 | 3.5351 | **0.002662** |
| LSVC - MLP | 0.98 | 0.8737 | 3.5351 | 0.8565 |
| LSVC - SVM | 0.3 | 0.8737 | 3.5351 | 0.9948 |

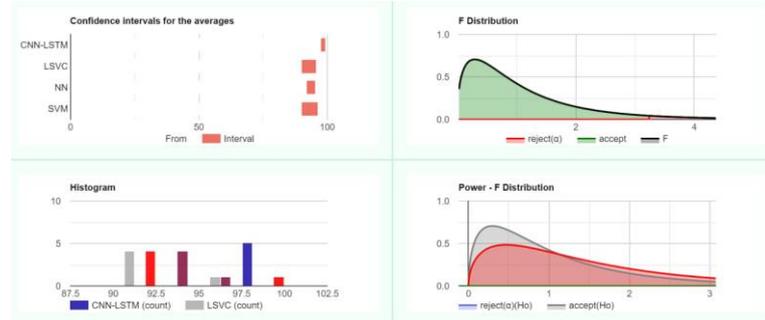

Figure 4. Statistical significance of CNN-LSTM compared to others

## V. CONCLUSION

In this experiment, we compiled an extensive set of medical radiological images to distinguish emergency patients from non-emergency patients. We introduced our CNN-LSTM model, specifically tailored with input parameters to address this challenge. The experiment comprises three main phases. In the initial phase, text data is extracted and prepared for preprocessing. Subsequently, in the second phase, the preprocessed data undergo word embedding for our deep learning model. The selected data is employed for training the model, while another portion is reserved for testing in our third phase. This process iterates over five cycles on our datasets with two and four classes in both balanced and imbalanced mode. The accuracy of our model is calculated, and for datasets with two labels, evaluation criteria are computed to assess the model's performance.

Finally, the experimental results are compared with the performance of ML models such as MNB, LR, LSVC, RF, MLP, DT, and SVM to demonstrate the effectiveness of our suggested model. The findings indicate that CNN-LSTM outperforms other ML models in all datasets, highlighting the superior accuracy of the DL algorithm compared to traditional ML models. In conclusion, the results of the analysis, incorporating two tests (one-way ANOVA and Tukey's HSD), affirm the robust performance of CNN-LSTM.

To advance and enhance future research endeavors, the following actions can be considered:
- Exploring solutions to further enhance the accuracy of the models.
- Investigating the performance of the models on larger datasets to assess scalability.
- Incorporating additional features into the dataset for a more comprehensive analysis